# MotàMot project: conversion of a French-Khmer published dictionary for building a multilingual lexical system

**Mathieu Mangeot**

GETALP-LIG Laboratory, 41 rue des mathématiques BP 53
F-38041 GRENOBLE CEDEX 9
France
Email: Mathieu.Mangeot@imag.fr

## Abstract

Economic issues related to the information processing techniques are very important. The development of such technologies is a major asset for developing countries like Cambodia and Laos, and emerging ones like Vietnam, Malaysia and Thailand.

The MotAMot project aims to computerize an under-resourced language: Khmer, spoken mainly in Cambodia. The main goal of the project is the development of a multilingual lexical system targeted for Khmer. The macrostructure is a pivot one with each word sense of each language linked to a pivot axi. The microstructure comes from a simplification of the explanatory and combinatory structure.

The lexical system has been initialized with data coming mainly from the conversion of the French-Khmer bilingual dictionary of Denis Richer from Word to XML format. The French part was completed with pronunciation and parts-of-speech coming from the FeM French-english-Malay dictionary. The Khmer headwords noted in IPA in the Richer dictionary were converted to Khmer writing with OpenFST, a finite state transducer tool.

The resulting resource is available online for lookup, editing, download and remote programming via a REST API on a Jibiki platform.

**Keywords**: French-Khmer dictionary, MotÀMot, Jibiki

## 1. Introduction

Economic issues related to the information processing techniques are very important. The development of such technologies is a major asset for developing countries like Cambodia and Laos, and emerging ones like Vietnam, Malaysia and Thailand.

The MotAMot project aims to computerize an under-resourced language: Khmer. This language is spoken mainly in Cambodia.

As indicated by V. Berment in his Ph.D. thesis (Berment, 2004), "With the development of personal computers and networks, the computer becomes now a tool for writing and communicating in the same way as paper and printing were before. Word processing, email, and even more advanced systems such as dictation or voice synthesis are widely used tools. The idea then is needed than appropriate information technologies tools must be added to traditional means without which the targeted goals can not be achieved any more". The computerization of a language such as Khmer occupies a central place in this wider context.

The main goal of the project is the development of a multilingual lexical system targeted for Khmer. The lexical system has been initialized with data coming mainly from the conversion of the French-Khmer bilingual dictionary of Denis Richer from Word to XML format.

## 2. Presentation of the resource to be built

### 2.1. Microstructure of the entries

The microstructure of the monolingual volumes is based on the Explanatory and Combinatorial Lexicography.

Each entry is based on the vocable. A vocable is either a group of lexies (word meaning) or an idiom.

Each entry consists of a headword and a pronunciation, followed by a list of word meaning (lexies). Each lexie is described either by a lexico-semantic formula (Melcuk & Polguère, 2006 ; Polguère, 2006) or a free gloss. For terms, it is possible to describe their domain. Then comes a translation link pointing to an axie (entry) of the pivot volume. It is followed by a list of examples and idiomatic expressions. Finally, a generic field is used to store additional information from reused dictionaries.

To cope with different contributor skill levels, the editing interface adapts itself and displays an appropriate granularity of information. For example, a novice contributor will be prompted for a simple gloss to characterize a lexical unit, while an expert linguist will describe a complete semantic formula. Similarly, only specific contributors have access to the list of lexical functions.

### 2.2. Macrostructure of the lexical database

The macrostructure is composed of a monolingual volume for each language and a central pivot volume. However, in order not to confuse users, they contribute through an interface with a classic view of a bilingual dictionary. Each bilingual link language A ➔ language B added via this interface (step 1 in Figure 1) is actually translated into the background by creating two interlingual links as well as an axie link representing the original translation to finally get: language A ➔ pivot axie ➔ language B (step 2 in Figure 1).

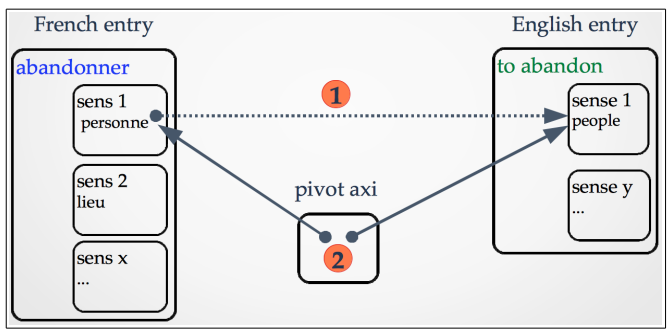

Figure 1: Building bilingual links

If a contributor wants to add a translation link between a vocable Va of language A and a vocable Vb of language B, s/he can establish this link at different levels. The ideal solution is to connect a word meaning (lexie) La of the vocable Va to another word meaning Lb of the vocable Vb. In this case, the link is bijective and Lb is also connected to La (point 1 in Figure 2).

If the vocable Vb has not yet defined specific word meanings or if the contributor can not choose between word meanings, s/he can connect directly the word meaning La to the vocable Vb. In this case, a new word meaning Lb is created with a draft level of quality and the link as well as the word meaning are marked for refinement (point 2 in Figure 2)..

In the case of reusing existing data, it is often impossible to link information to a specific word meaning. In this case, we add to the end of the vocable Va the information that one of the meanings of vocable Va can be connected to a meaning of vocable Vb but this information will not be added to Vb (point 3 in Figure 2). It will of course be marked as to be refined with emergency!

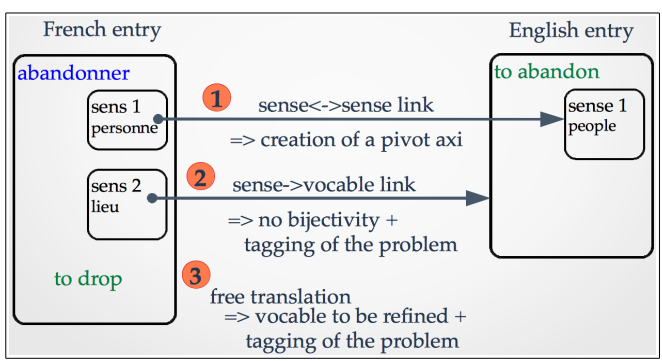

Figure 2: Different types of links

With the pivot macrostructure, if two links language A ➔ language B and language B ➔ language C exist, then it will automatically create a link language A ➔ language C whose level will be marked as a draft and to be revised.

## 2.3. Levels for contributions and contributors

Each information of each entry is assigned a quality level. The levels range from 1 star for a draft (recovered data whose quality is not known) to 5 stars, quality certified by an expert (e.g.: a link translation validated by a sworn translator).

Similarly, contributors will be assigned a skill level (1 to 5 stars as well). 1 star is the level of an unknown novice in the community and 5 stars is the level of a recognized expert.

Then, when a level 3 contributor revises a level 2 entry, the entry automatically goes up to level 3. Similarly, if the work of a contributor is systematically validated without corrections by other contributors to the next level, it can automatically switch to the next level after a certain threshold (e.g. 10 contributions).

To go further, we plan to analyse the work of the contributors. If a person contributes massively eg on a particular area, the system will automatically send regular proposals for contribution in the domain.

# 3. Preparation of existing data

In order to encourage contributions, it is preferable to provide a skeleton dictionary to modify later rather than an empty dictionary. For each language involved, a list of words of this language will be recovered to create an initial list of entries. It will always be possible to create a new article, but the creations will be subject to verification.

## 3.1. Conversion of a French-Khmer dictionary into XML

### 3.1.1. Description of the dictionary

For the Khmer language, there is at present a French-Khmer dictionary (Richer et al., 2007), which started in the late 90s, was completed in 2006 by a small group of computer scientists gathered in the "Pays Perdu" NPO created by Denis Richer, french ethnolinguist established in Siem Reap (Cambodia). This first version of the dictionary was published in spring 2007 and has 13,249 entries. Each entry is composed by a French headword, in some cases a part-of-speech and a list of word senses. Each word sense has a gloss in French and a translation in Khmer. The dictionary was originally encoded in Word format. An example of the original file can bee seen on Figure 3.

| | |
|---|---|
| abondant, e (fruits, riz...) | (dɑ̃el) sɑmbŏ̄ |
| — (pluie) (trempé-humide) | (dɑ̃el) cŏ̝k-coam |
| abonnement (magazine) | cĭə̄w-prŭ̄ cam |
| — (téléphone) | kă̄ː bɑŋ-sĕ̄va̋ |
| abonner (sur pied-nom (s'inscrire)-commercer) | coh-chmŭəh-cĭə̄w |
| abonner (s') (à un magazine) | cĭə̄w-prŭ̄ cam |
| | bɑŋ-sĕ̄va̋ |
| abord (adv.) (d'—) | mun-dɑmbŏ̄ŋ / dă̄əm-lă̄əj |
| aborder (accoster) | cŏ̄l-cŭ̄ t / cŏ̄l-cɑt |
| — (qqn) (appeler-arrêter) | haw-bɑŋchop |
| — (commencer-discuter-sur-sujet-un) | phdă̄əm-cŏ̄ʔcĕ̄k-ɑmpĭ̄ pɑɲə-hă̄ː mŭ̄əj |
| aboutir (arriver à destination, déboucher) | tŏ̄w-dɑl |
| — (avoir-résultat) | mĭ̄ən-lŏ̄ttəphŭ̄ l |
| — (devenir) (aller-être) | tŏ̄w-cĭ̄ə |

Figure 3: Extract of the dictionary in Word format

### 3.1.2. Conversion of the file into XML

We first converted the dictionary from Word format to a lexical resource in XML format according to the methodology described in (Mangeot & Enguehard, 2013). The Open Document Format has the great advantage of being based on XML. Instead of a conversion, the contents of the XML document has been retrieved, then transformed to obtain what we want.

A document in ODF format is actually a zip archive containing multiple files including the text content in XML. This content is stored in the "content.xml" file in the archive. To retrieve this file, just follow some clever manipulations. On MacOs, create an empty folder and then copy the .odt file inside. Then, open a terminal and run the "unzip" command to unzip the file. On Windows, you must change the .odt file extension into .zip and then open the. zip archive.

The file "content.xml" can now be extracted from the archive and then renamed and placed in another location. It becomes the base file on which we will continue our work. The next step consists in editing this file with a "raw" text editor with syntax highlighting and regular expressions based search and replace.

One may first think that since the source file "content.xml" is already in XML, it may be enough to write an XSLT stylesheet to convert the file into an XML dictionary, but the XML used in the source file is completely different from the XML targeted. Indeed, the source file comes from a word processor. It is designed for styling a document and not for structuring a dictionary entry. Therefore, it is finally easier to convert the XML file "by hands" with regular expressions than to write an XSLT stylesheet for automatically converting the source file.

### 3.1.3. Tagging each part of information

All pieces of information were explicitly tagged by replacing the ODF markup by a new "homemade" tagset. Each piece of information is usually distinguished from others in the original file with a different style. Search/replace operations were performed for each type of information.

The original dictionary file has tow columns. The French is on the left one and the Khmer on the right one. Each entry has one or several word senses. Each word sense is written on one line and has a khmer translation on the right column. Therefore, each line was tagged with "sens" tag (see Figure 4).

```
<article>
 <vedette>abondant</vedette>
 <sens>
  <glose>abondant, e (fruits, riz...)</glose>
  <traduction><api>(dɑ̃el) sɑmbŏ̄</api>
  </traduction></sens>
 <sens>
  <glose>(pluie) (trempé-humide)</glose>
  <traduction><api>(dɑ̃el) cŏ̝k-coam</api>
  </traduction></sens>
</article>
```

Figure 4: French entry "abondant" after XML conversion

On the original Word file, feminine forms of adjectives are separated from the masculine with a comma. The feminine form were separated in order to clearly identify the headword of each entry.

The original form of the entry is duplicated in the gloss field.

The XML file obtained is a unique volume containing French entries translated into Khmer.

## 3.2. Restructuring and links reification

### 3.2.1. Conversion into the MotÀMot structure

The XML data obtained is then converted into the structure chosen for the MotÀMot project.

In files coming from word processors, the data structure is usually implied. New structural elements must be added to move towards a more standardized structure. Concerning standards, Lexical Markup Framework (LMF) (Romary et al., 2004) became an ISO standard in November 2008 (Francopoulo et al., 2009). It suits ideally our goals. As it is a meta-model and not a format, the principle of the LMF model can be applied to the entry structure and keeping our tags without using the LMF syntax (Enguehard & Mangeot, 2013). The entry tag corresponds to the LexicalEntry object; the head tag corresponds to the Form object; The headword tag corresponds to the Lemma object; the sense tag corresponds to the Sense object; The gloss tag corresponds to the Definition object; the refaxie tag corresponds to the Equivalent object (Figure 6).

The core meta-model LMF is shown in Figure 5.

Unique identifiers are also added to entries and word senses (Figure 6). They will be used afterwards for linking entries.

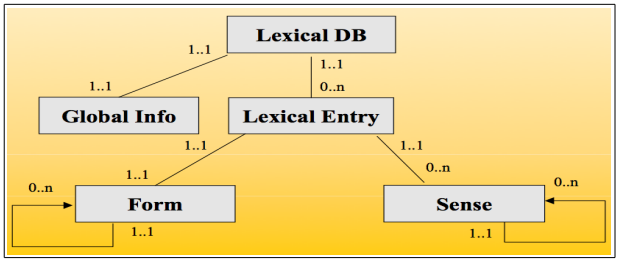

Figure 5: LMF core metamodel

```
<m:entry id="fra.abondant.27.e" level="">
 <m:head>
  <m:headword>abondant</m:headword>
  <m:pos></m:pos>
 </m:head>
 <m:sense id="s1" level="">
  <m:gloss>abondant, e (fruits, riz…)</m:gloss>
  <m:translations>
   <m:translation>sambō</m:translation>
  </m:translations>
 </m:sense>

  <m:sense id="s2" level="">
   <m:gloss>(pluie) (trempé-humide)</m:gloss>
   <m:translations>
    <m:translation>cɔ̆k-coam</m:translation>
   </m:translations>
  </m:sense>
 </m:entry>
```

Figure 6: French entry "abondant" after restructuring

### 3.2.2. Link reification

We then converted this unique volume into three volumes : one French, one pivot and one Khmer by eliciting the translation links. From each French->Khmer translation, we obtain 2 links : one French->Pivot and one Pivot->Khmer.

The entry `fra.abondant.27.e` has 2 Khmer translations: "sambō" & "cɔ̆k-coam".

They were elicited into the 4 following links:

1. Sense `s1` of French entry `fra.abondant.27.e` linked to Axi entry `axi.[fra:abondant,khm:sambō].27.1.e`;

2. Axi entry `axi.[fra:abondant,khm:sambō].27.1.e` linked to Khmer entry `sambō`;

3. Sense s2 of French entry `fra.abondant.27.e` linked to Axi entry `axi.[fra:abondant,khm:cɔ̆k-coam].27.2.e`;

4. Axi entry `axi.[fra:abondant,khm:cɔ̆k-coam].27.2.e` linked to Khmer entry `cɔ̆k-coam`.

Figure 7 shows a French entry after reification.
The axi volume is obtained by gathering all the axies in the same file.

The Khmer volume is obtained by gathering all the Khmer entries and by merging them when they have the same Khmer word as headword (at this stage, the comparison is based on the IPA of the Khmer word). Each link to an axie constitutes a sense.

The volume will then be sorted by Unicode sort order of the Khmer headword in Khmer writing after the process described in part 4.

The French volume contains 13,249 entries; the Khmer volume contains 23,766 khmer entries; the axie volume contains 32,402 axies linking to both a French and a Khmer entry.

### 3.2.3. Adding data in the French volume

The French entries were enriched with data from the FeM French-English-Malay dictionary : the pronunciation of the headword and the part-of-speech for the non-homonymous vocables.

```
<m:entry id="fra.abondant.27.e" level="">
   <m:head>
    <m:headword>abondant</m:headword>
    <m:pronunciation>ABON-DAN-</m:pronunciation>
    <m:pos>adj.</m:pos>
    <m:fem_form>abondante</m:fem_form>
    <m:fem_pron>ABON-DAN-T</m:fem_pron>
   </m:head>
  <m:sense id="s1" level="">
   <m:gloss>abondant, e (fruits, riz…)</m:gloss>
   <m:refaxie idrefaxie="axi.[fra:abondant,
khm:sambō].27.1.e" />
  </m:sense>

  <m:sense id="s2" level="">
    <m:gloss>(pluie) (trempé-humide)</m:gloss>
    <m:refaxie idrefaxie="axi.[fra:abondant,
khm:cɔ̆k-coam].27.2.e" />
   </m:sense>
  </m:entry>
```

Figure 7: French entry "abondant" after reification

## 4. Conversion of the IPA in Khmer alphabet

The Khmer part is a simple phonetic transcription of the Khmer translations written in a special API font (SILSophia IPA93) created by the Summer Institute of Linguistics. We developed a program of transcription from the phonetic form to the Khmer writing[1] based on a finite state transducer (FST): OpenFst[2]. This program is available online for use at the following URL:
http://jibiki.univ-savoie.fr/khmer/

The source code is also available for download from the same page. This work is licensed under the Creative Commons Zero license, public domain.

### 4.1. IPA normalisation

The first step is to normalize the IPA notation. It consists in two different parts. The first part is a correction of mistakes in the IPA notation. Some IPA letters that do not

---

exist in Khmer pronunciation were replaced by the correct ones.

The second part is the normalisation of combined letters. In Unicode, some letters with diacritics can be written by combining a basic character with different marks (combining macron, ring below, etc.) but also exist in one global character. When one character exists, the combining characters where replaced by the equivalent global character.

| Original | Replacement |
|---|---|
| Corrections | |
| y | j |
| f | hv |
| n | ŋ |
| Normalisations | |
| a + ¯ | ā |
| o + ¯ + ₒ | ǭ |

Table 1: replacement operations for API normalisation

## 4.2. Intermediate notation

During this step, two operations are performed at the same time.

The first one consists in grouping the series of API letters that correspond to one letter in Khmer.

The longest matches are processed before.

The second one consists in tagging the type of syllable. In Khmer, each consonant can be written in two different ways following the composition of the syllable in which the consonant is included. These two variants are considered as two different letters with different Unicode codes. We call the two variants A and B and we tag each consonant with its type.

For tagging the type of consonants, several heuristics are applied in the following order:

In the original file, Khmer words noted in IPA are separated by the '-' character. This feature was used in the conversion when word beginning and endings require special treatments. Some vowels beginning a word like "a" and "ə" are specific Khmer letters.

| - a | - a |
|---|---|
| - ə t | - ət t |
| - ā | -a ā |

The consonant "l" ending a word is a type B consonant.

| l - | l B - |
|---|---|

There are three types of vowels:
- some are located after a type A consonant;

| ā e | A āe |
|---|---|

- other after a type B consonant;

| u ə | B uə |
|---|---|
| ō a | B ōa |

- the third category can be located indifferently after the two types of consonants. Hopefully, only two vowels are in this case.

| ū ə | ūə |
|---|---|
| ɯ̄ ə | ɯ̄ə |

After the three and two IPA letter combinations, come the unique letters:

| e | A e |
|---|---|
| ū | B ū |

Next, the consonants are grouped:

| c h | ch |
|---|---|
| k h | kh |
| p h | ph |
| t h | th |

For some consonants followed by third type vowels, it is possible to attribute them automatically a type.

| b ūə | b A ūə |
|---|---|
| ch ūə | ch B ūə |
| d ūə | d A ūə |
| t ūə | t ūə |
| b ɯ̄ə | b B ɯ̄ə |
| k ūə | k B ūə |
| t ɯ̄ə | t B ɯ̄ə |

## 4.3. Khmer generation

The third step consists in generating Khmer writing from the intermediate notation with the two types of consonants.

The vowels are replaced with the same Khmer letters:

| ā | ា |
|---|---|
| ɯ̄ə | ើ |

The consonants are replaced following their type:

| kh + v + A | ខ្ញុំ |
|---|---|
| kh + v + B | ឈ្មោះ |

The Khmer language writing system has many ambiguities,. Some letters (most of them word endings) are not pronounced. The result is far from perfect. Nevertheless, it was still added to entries and will then be reviewed by khmer native speakers.

## 5. Availability on the Web with the Jibiki platform

Jibiki is a generic platform for manipulating online lexical resources (Mangeot, 2001) with users and group management, consultation of heterogeneous resources and generic dictionary articles edition. This is a community website originally developed for the Papillon project (Mangeot et al., 2003).

Jibiki distinguishes form other lexical resources development platforms such as TshwaneLex from tshwanedje.com, the Dictionary Publishing System from IDM or FieldWork from the SIL on the following points:

- it is completely free and open-source;
- it allows to browse and edit any kind of XML lexical resource without modifying the XML structure or tags (so far, more than 100 resources were successfully imported into a Jibiki platform);
- it allows to manipulate complex macrostructures such as pivot ones or lexical networks;
- it is available online from a simple Web browser;
- it is remotely programmable via a REST API.

### 5.1. Description of the platform

The platform is programmed entirely in Java based on the environment "Enhydra". All data is stored in XML format in a database (Postgres). This website mainly offers two services: a unified interface for simultaneous access to many heterogeneous resources (monolingual, bilingual dictionaries multilingual databases, etc...) and a specific editing interface to contribute directly to the dictionaries available on the platform.

Several lexical resources construction projects used or still use this platform successfully (Mangeot & Chalvin 2006; Sérasset et al., 2006). The source code for this platform is published under the LGPL open-source licence and available for free download from the open forge of the LIG laboratory[3].

An instance of the platform has been adapted specifically to the MotÀMot project. The language of the interface is French. It is also envisaged to localize it in Khmer.

### 5.2. Visualisation of Khmer text in browsers

Unfortunately, widespread web browsers based on the Webkit engine such as Safari or Chrome but also word processors such as OpenOffice family are still not able to display correctly all the Khmer letters. Some letters have to be composed. The order must change. In the following example, the last letter of the word "jɔ̄n-hɔh" (plane) has a dash circle which is in fact an empty placeholder for a vowel. On the same system with the same configuration (MacOs X 10.6.8), Firefox is able to fill in the placeholder (Figure 9) but not Safari (Figure 8).

យអណហាៈ /jɔ̄n-hɔh/

Figure 8: Khmer word not correctly displayed in Safari

យអណហា: /jɔ̄n-hɔh/

Figure 9: Khmer word correctly displayed in Firefox

### 5.3. Volume Lookup

Two different lookup interfaces are available to the user. The volume lookup allows the user to lookup a word or prefix on a specific volume, French or Khmer. The Khmer entries can be searched from their headword in IPA or Khmer writing. On the left part of the result window, the volume headwords are displayed, sorted in alphabetical order. An infinite scroll allows the user to browse the entire volume. On the right part of the window, the entries previously selected on the left part are displayed (see Figure 10).

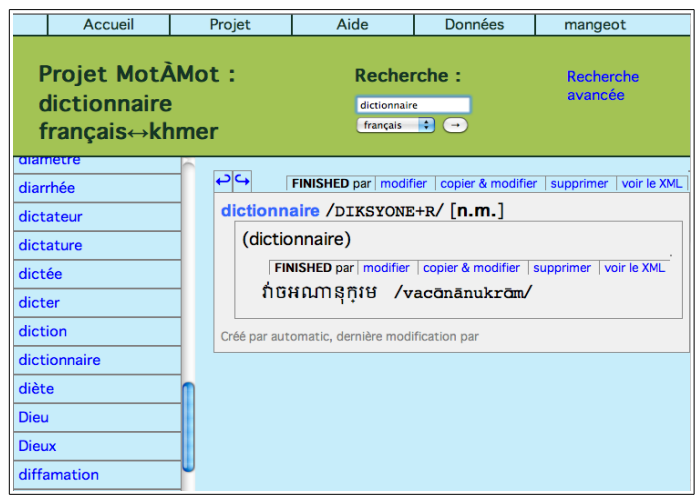

Figure 10 : Entry "dictionnaire" of the French-Khmer dictionary in Jibiki

### 5.4. Advanced lookup

The advanced lookup is available for complex multi-criteria queries. For example, it is possible to lookup an entry with a specific part-of-speech, and created by a specific author. On the left part of the result window, the headwords of the matching entries are displayed, sorted in alphabetical order. An infinite scroll allows the user to browse all the matching entries. On the right part, the entries previously selected on the left part are displayed.

## 5.5. Online editing

The editing module is based on an HTML interface model instantiated with the entry to be edited (see Figure 11). The model can be generated automatically from a description of the structure of the entry using an XML schema. It can then be modified to improve the screen rendering. The only information required to publish a dictionary article is the XML schema representing the structure of this entry. In addition, it is possible to edit any type of dictionary provided that it is encoded in XML. Figure 8 shows the editing interface for the French entry "dictionnaire". The link to the pivot axie is indicated in the `refaxie` field of the `sense` block. If a contributor wants to add a new link to a Khmer entry, s/he will use the curved arrow button on the left of the `reflexie` field in the `translation` block. Then s/he will be able to look up a word in the Khmer volume. When a Khmer entry is selected, an axie is generated in the background and the appropriate links are added to the French and Khmer entries. The resulting axie link is then shown in the `refaxie` field of the `sense` block.

Figure 11: Editing interface of the French entry "dictionnaire"

## 5.6. XML data available for download

The XML data of the dictionary (French, Khmer and axies volumes) is available for download with a CreativeCommons CC by license via the "Données" tab on the MotÀMot website. Data exports from the database will be regularly performed and the new versions available for download.

## 5.7. Remote access via a REST API

Once dictionaries are uploaded into the Jibiki server, they can be accessed via a REST API[4]. Lookup commands are available for querying indexed information: headword, pronunciation, part-of-speech, domain, example, idiom, translation, etc. (see Figure 12). The API can also be used for editing entries provided that the user previously registered in the website.

---

4 http://jibiki.univ-savoie.fr/motamot/Api.po

Figure 12: API documentation for querying entries

## 6. Conclusion

The MotÀMot project has now ended. The project website is available at the following URL:
http://jibiki.univ-savoie.fr/motamot/
Users can lookup and edit French and Khmer dictionary entries. They can also download the resulting XML data. The platform can be used remotely from other applications via a REST API.
The quality of the data can be increased subsequently. Particularly the Khmer entries and headwords that were generated automatically need revision by Khmer native speakers. Of course, people can already voluntarily contribute, but it is very difficult to do so without at least a community manager that can help and motivate contributors. We are now looking for funds in order to found a second project focused on the data revision.

## 7. Acknowledgements

The MotÀMot project has been partly funded by the Agence Universitaire de la Francophonie. We also thank Vanra IENG for his work on the project.